\pdfoutput=1

\documentclass[11pt]{article}

\usepackage[final]{acl}

\usepackage{times}
\usepackage{latexsym}

\usepackage[T1]{fontenc}

\usepackage[utf8]{inputenc}

\usepackage{microtype}
\usepackage{newfloat}
\usepackage{listings}
\usepackage{algorithm}
\usepackage{algorithmic}
\usepackage{hyperref}
\usepackage{tabularx}
\usepackage{cite}
\usepackage{natbib}  
\usepackage{caption} 
\usepackage{xcolor}
\usepackage{graphicx}
\usepackage{courier}
\newfloat{listing}{tb}{lst}{}
\floatname{listing}{Listing}
\title{ConvNLP: Image-based AI Text Detection}

\author{Suriya Prakash Jambunathan  \\ New York University \\ sj3828@nyu.edu
        \\\And 
        \hspace{1cm}
          Ashwath Shankarnarayan \\ \hspace{1cm} New York University \\ \hspace{1cm} as16494@nyu.edu
         \\\And
        Parijat Dube \\ IBM Research \\ pdube@us.ibm.com
         }

\begin{document}
\maketitle
\begin{abstract}
The potentials of Generative-AI technologies like Large Language models (LLMs) to revolutionize education are undermined by ethical considerations around their misuse which worsens the problem of academic dishonesty. LLMs like GPT-4 and Llama 2 are becoming increasingly powerful in generating sophisticated content and answering questions, from writing academic essays to solving complex math problems. Students are relying on these LLMs to complete their assignments and thus compromising academic integrity. Solutions to detect LLM-generated text are compute-intensive and often lack generalization. This paper presents a novel approach for detecting LLM-generated AI-text using a visual representation of word embedding. We have formulated a novel Convolutional Neural Network called ZigZag ResNet, as well as a scheduler for improving generalization, named ZigZag Scheduler.
Through extensive evaluation using datasets of text generated by six different state-of-the-art LLMs, our model demonstrates strong intra-domain and inter-domain generalization capabilities. 
Our best model detects AI-generated text with an impressive average detection rate (over inter- and intra-domain test data) of 88.35\%. Through an exhaustive ablation study, our ZigZag ResNet and ZigZag Scheduler provide a performance improvement of nearly 4\% over the vanilla ResNet. The end-to-end inference latency of our model is below 2.5ms per sentence.
Our solution offers a lightweight, computationally efficient, and faster alternative to existing tools for AI-generated text detection, with better generalization performance. It can help academic institutions in their fight against the misuse of LLMs in academic settings. 
Through this work, we aim to contribute to safeguarding the principles of academic integrity and ensuring the trustworthiness of student work in the era of advanced LLMs. 

\end{abstract}

\section{Introduction}
Integration of Generative AI technologies like Large Language Models (LLMs) with education has huge potentials to enhance teaching quality and effectiveness. They can be used to generate personalized learning content, develop instruction support tools, evaluate student assignments, and generate rich content on a variety of topics~\citep{gan2023large}. At the same time, there are  ethical issues regarding their use in academic settings~\citep{becker2023} as LLMs can contribute to misinformation and plagiarism~\citep{khalil2023}. Since LLMs are trained on corpora of human written text, they are capable of generating sophisticated text which is almost indistinguishable from human-authored content. It is extremely challenging to distinguish between human and AI-generated texts, leading to an increased risk of plagiarism and a decrease in the integrity of academic contributions. The problem is further aggravated with the advent of newer and more powerful LLMs like OpenAI's GPT-4 \citep{openai2024gpt4} and LLAMA 2 \citep{touvron2023llama}  models. Therefore, it’s imperative to develop robust methods to identify AI-generated text within the academic realm to preserve the authenticity and value of student's work.

In this paper, we suggest a novel method to detect AI-generated text using a vision model. This method involves feeding image representations of the generated word embedding to the deep neural network. This approach helps in combating the misuse of AI for academic dishonesty by efficiently distinguishing between human and machine-generated texts. Moreover, by studying the qualities and patterns of AI-generated text, this research aims to deepen our understanding of these advanced language models.

State-of-the-art NLP models offer a wide range of applications, including the capability to detect AI-generated text. However, they come with significant limitations \citep{brown2020language}, primarily due to their need for vast amounts of data and computing power. According to a report by OpenAI, training their GPT-3 model required 175 billion parameters and over 3 million GPU hours \citep{brown2020language}. Today, several models can analyze and classify the text using NLP. However as mentioned above, NLP text classification models are computationally expensive for training and inference. Similar to LLMs, they require heavy GPUs, longer training time, huge and diverse collection of data, etc.

In contrast, image processing models, which have been evolving over time, present fewer limitations. Vision models use significantly less number of parameters \citep{villalobos2022machine}. Also, these models have improved over time and show better and more consistent detection rate scores. Such models need less training time and are less GPU-intensive than text-based transformer models. However, one main challenge with solely using conventional vision models like CNNs is their lack of textual context. This means that while they are efficient in processing images, they may not fully grasp the nuanced information conveyed through text.

To avoid such limitations, we formulated a Text-to-Image embedding, which bridges language and vision domains. Using Universal Sentence Encoder~\citep{cer2018universal}  we can preserve the spatial and linguistic relation between sentences. Then using visual representation of the embedding, we can efficiently detect AI-generated content with the help of vision models.

Major Contributions in this paper are:
\begin{itemize}
  \item ZigZag ResNet, a novel variant of ResNet
  \item ZigZag Scheduler, a novel scheduler aimed at improving generalization
  \item Image-based embedding approach using Universal Sentence Encoder
  \item Generalization study of our proposed approach across six different state-of-the-art large language models
  \item Ablation study of the proposed model and scheduler
\end{itemize}

Section~\ref{related} deals with related work. Our methodology and model are explained in Section~\ref{methodology}. Section~\ref{data} provides details of the different data sources used in our model evaluation and the pre-processing steps. Experimental details are in Section~\ref{experiment} with a discussion of our results and observations. Scaling of our model's inference time over CPU and GPU with increasing batch sizes and sentence lengths is studied in Section~\ref{scaling} Finally, we conclude in Section~\ref{conclusion}.  

\section{Related Work}
\label{related}
Recent advancements in Generative AI, particularly in Large Language Models (LLMs), have escalated concerns regarding the authenticity of digital content. The body of research focusing on differentiating AI-generated text from human-authored content has grown significantly. 



\citet{Alghamdi2022} provides a comprehensive comparison between machine learning and deep learning techniques for detecting fake news. Their study emphasizes the effectiveness of deep learning models in identifying nuanced patterns indicative of misinformation, a technique that can be paralleled in the detection of AI-generated texts, considering both domains require discerning subtle inconsistencies in content.



%

\citet{abdali2024decoding} provides a detailed survey of existing methods for AI generated text detection and categorizes them as supervised~\citep{bakhtin2019real,li2023deepfake,quidwai-etal-2023-beyond}, zero-shot~\citep{su2023detectllm,mitchell2023}, retrieval-based~\citep{krishna2023paraphrasing,Liang2023}, watermarking~\citep{kirchenbauer2023reliability,yang2023watermarking}, and feature~\citep{yang2023dnagpt,yu2023gpt} detection based.

Existing techniques for AI-generated text detection are fragile and susceptible to adversarial attacks involving paraphrasing~\citep{krishna2023paraphrasing} and spoofing~\citep{pang2024attacking}. With careful prompt engineering, LLM-generated text can be made increasingly harder to detect~\citep{lu2023large}.

Several tools are available for AI-generated text detection, including GPTZero~\citep{gptzero}, DetectGPT~\citep{detectgpt-tool}, Turnitin~\citep{turnitin}, CopyLeaks AI Content Detector~\citep{copyleaks}, and Plagium Originality AI Detector~\citep{plagium}. 
Due to the continual introduction of new LLMs, existing tools, trained on data generated using older language models, fail to generalize and have poor performance in detecting text generated using newer LLMs. 
An earlier tool by OpenAI for AI text detection was discontinued due to its poor performance~\citep{openai23} within six months after its release.
Additionally, LLMs are exposed to indirect data leaking~\citep{balloccu-etal-2024-leak}  and are iteratively improving their content generation capability using user interaction data. 

Due to the black-box nature of commercially available AI-generated text detection tools, the predictions made by such tools lack explainability which limits their usage in academic settings where false positives can lead to unfair academic assessments. Fairness concerns of AI text detectors in educational settings were the focus in ~\citet{Liang2023} which found current text detectors often raise false alarms for non-native English writing.

\citet{weber-wulff2023} evaluated the suitability and effectiveness of popular tools (including Turnitin and PlagiarismCheck~\citep{plagiarismcheck}) for plagiarism detection in academic settings. The authors concluded that these tools are inaccurate, unreliable, biased, and can be fooled by content obfuscation techniques like machine translation, patchwriting, and paraphrasing. 

 Our proposed approach is a supervised detection, however, unlike existing techniques, our solution only uses LLM for generating word embeddings while the classifier is CNN-based. Efforts in combining different domains, like utilizing image processing techniques for text analysis, have been explored earlier by~\citep{butnaru2017image,wang2017liar} showcasing the effectiveness of hybrid models for text classification. 

\section{Methodology}
\label{methodology}

\begin{figure*}[t]
\centering
\includegraphics[scale=0.5]{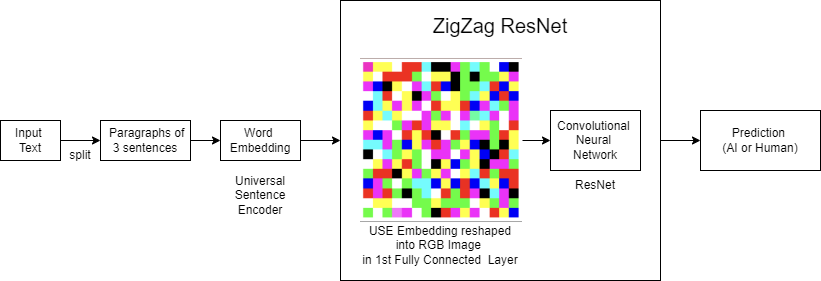}
\caption{Our methodology for AI-generated text detection}
\label{fig:methodology}
\end{figure*}

Our novel approach involves utilizing a vision model for classifying text, which is usually done using transformer networks. Figure~\ref{fig:methodology} shows our methodology which involves generating word embeddings for the input text, converting the word embeddings into a colored (RGB) image, and feeding this visual representation to a convolutional neural network-based image classifier. 

\begin{figure*}[t]
\centering
\includegraphics[width=\textwidth]{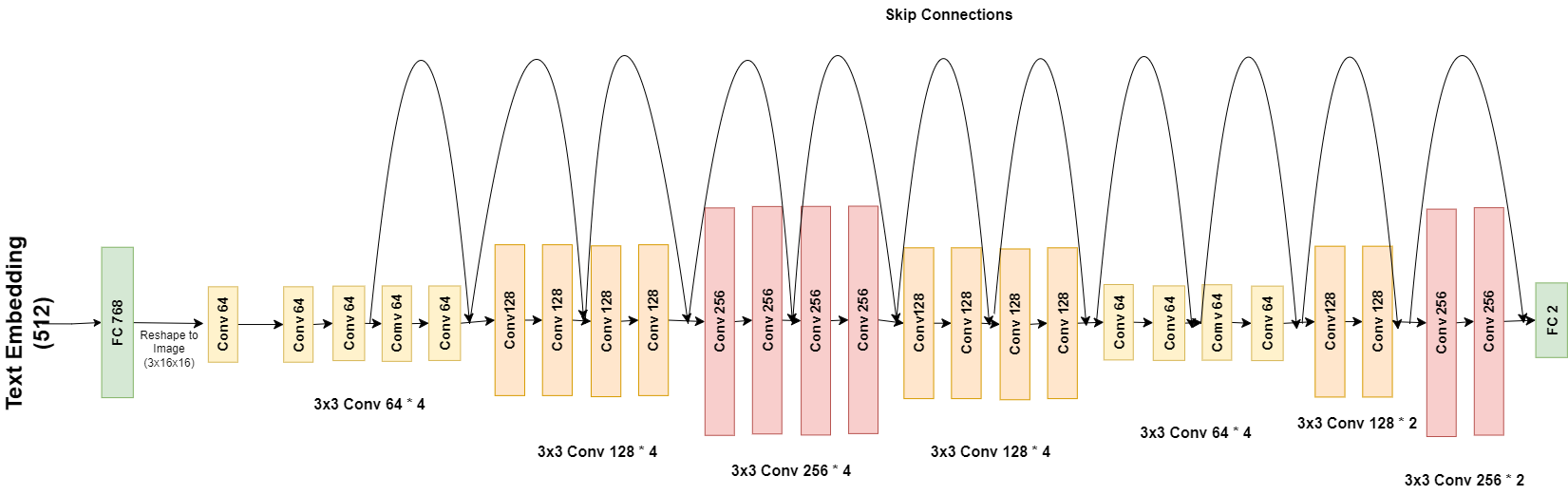} 
\caption{ZigZag ResNet Architecture used in our solution}
\label{fig3}
\end{figure*}

Our model incorporates the principles of residual architecture associated with vision tasks, to classify text in a novel manner as shown in Figure~\ref{fig3}. ResNets~\citep{he2015deep} is one of the most popular, parameter-efficient, and high-performing models for image classification. The regular ResNet model, including 18-layer ResNet~\citep{he2015deep}, has a uniform channel size progression. Optimizing this architecture by introducing a variety of channel sizes was identified as an effective strategy to enhance its performance without increasing the model's complexity. We are utilizing the Universal Sentence Encoder (USE) to generate a fixed-length embedding. The USE word embedding array of size 512 is fed to a fully connected layer of size 768 whose output is reshaped to a 3x16x16 matrix emulating an RGB as shown in Figure~\ref{fig:methodology}. This image is passed to a sequence of convolutional layers. The channel sizes of these convolutional layers gradually fluctuate between the range 64 to 256 multiple times. This distinctive neural network configuration symbolizes a zig-zag movement, hence the name \texttt{ZigZag ResNet}. The model introduces more variety and utilizes several distinct ResNet blocks, in contrast to four unique blocks in the original ResNet-18.

The input text is reorganized into paragraphs of three sentences each. Then, USE is used to encode the three sentences into a floating-point array of size 512. To emulate image pixels, the embedding values are mapped to the range 0 to 255. The mapped embedding array is passed to the \texttt{ZigZag ResNet}. The first fully connected layer in the model converts the embedding into an image. This image is processed through the \texttt{ZigZag ResNet} architecture, which incorporates the diversified channel size progression. Through training, the model learns to associate certain spatial patterns in the embedding image with AI or human characteristics.

We devised a custom learning rate scheduler. Contrary to regular schedulers which only reduce the learning rate with increasing epochs, our \texttt{ZigZagLROnPlateauRestarts} Scheduler (ref. Algorithm~\ref{alg:algorithm}) increases or decreases the learning rate based on the performance of the model on the training and validation data. If the custom-defined performance metric increases, the learning rate will be increased. Similarly, the learning rate will be decreased if the performance metric decreases. This is a way of penalizing or rewarding the model based on the performance it provides. To prevent exploding or diminishing the learning rate, we will be resetting the learning rate after a fixed number of epochs. This mechanism is similar to the Warm Restarts scheduler \citep{loshchilov2017sgdr}. With the custom learning rate scheduler, we are able to achieve slight performance improvement in detection rate. 

\subsection{Hyper-Parameter Tuning}
We tuned our \texttt{ZigZag ResNet} model with HC3 dataset \citep{HC32023} to determine the best hyper-parameter configuration. HC3 is a thoroughly curated human-ChatGPT comparison corpus introduced in~\citet{guo2023close}. We also tried out different types of tokenizers, including the GPT tokenizer \citep{noauthororeditor}, BART tokenizer \citep{lewis2019bart}, Parts of Speech tagging \citep{schmid1994partofspeech}, and BERT \citep{devlin2019bert} to evaluate their impact on model performance. We chose the BERT tokenizer as it is lightweight compared to other LLMs. Furthermore, we experimented with varying lengths of text inputs to assess the model's performance under different conditions. This ranged from processing single sentences to analyzing inputs consisting of up to five sentences. Along with tuning the optimizer, we also tuned our \texttt{ZigZagLROnPlateauRestarts} scheduler (ref. Algorithm~\ref{alg:algorithm}). The hyperparameter values used in our evaluation are shown below.

\begin{listing}[h]
\textbf{Hyper-Parameters and Loss Function:}

Number of Parameters: 5,283,266\\
Batch Size: 32 \\
Optimizer: SGD \\
Learning Rate: 0.001, momentum = 0.8\\
weight decay = 0.005, Nesterov \\
Loss Function: CrossEntropy Loss \\
Scheduler: ZigZagLROnPlateauRestarts\\
mode = max, LR = 0.001 up-factor = 0.3, \\down-factor = 0.5, up-patience = 1, \\down-patience = 1, restart-after = 30
\end{listing}

\begin{algorithm}[t]
\caption{\texttt{ZigZagLROnPlateauRestarts}}
\label{alg:algorithm}
\textbf{Input}: Performance Metric, current learning rate\\
\textbf{Parameter}: mode, up factor, down factor, up patience, down patience, restart after\\
\textbf{Output}: updated learning rate\\
\begin{algorithmic}[1] 
\STATE Let Current Learning Rate = lr 
\STATE Let Previous Metric = prev metric
\STATE {num epochs ++}
\IF {(mode == min and metric $<$ prev metric) or (mode == max and metric $>$ prev metric)}
    \STATE {best lr = lr}
    \STATE {num bad epochs = 0}
    \STATE {num good epochs ++}
    \IF {num good epochs $>$ up patience}
        \STATE new lr = lr * (1 + up factor)
        \STATE num good epochs = 0
    \ENDIF
\ELSE
    \STATE {num bad epochs ++}
    \STATE {num good epochs = 0}
    \IF {num bad epochs $>$ down patience}
        \STATE new lr = lr * (1 - down factor)
        \STATE num bad epochs = 0
    \ENDIF
\ENDIF
\STATE {prev metric = metric}
\IF {num epochs \% restart after == 0}
    \STATE {lr = best lr}
\ENDIF
\end{algorithmic}
\end{algorithm}

\section{Data}
\label{data}
We evaluated our methodology on four different datasets (Wiki-Intro, HC3, Alpaca-GPT4, DAIGT) consisting of text generated by six different state-of-the-art LLMs.
 The Wiki-Intro Dataset \citep{aaditya_bhat_2023} is generated by prompting ChatGPT to complete the introductory paragraph of Wikipedia, given the first few words. The HC3 Dataset \citep{HC32023} is generated by prompting publicly available multi-domain questions to ChatGPT. The Alpaca-GPT4 Dataset \citep{peng2023instruction} is generated by prompting general questions from various domains to GPT-4. The DAIGT Dataset \citep{DAIGT-V2} is generated by prompting multi-domain essays to multiple large language models: Mistral, Claude, Llama and Falcon. The AI-generated samples in all of the above datasets are created by prompting these questions to the respective generative AI models and storing the responses as the generated samples. All four of the datasets mentioned above contain human-generated samples as well. After doing our preprocessing step, we have around 1.3 million human-written text samples. Table~\ref{tab:LLM} details the parameter counts of the LLMs and the number of data samples we have acquired for our training process. Due to the varied number of data samples for each LLM in the dataset, we randomly sampled that number of samples from the human-generated text dataset to use a balanced dataset for training. Therefore, our individual training datasets from LLMs contain equal amounts of AI-generated and human-written samples. However, the human-written content in each dataset is different due to the disparity in the availability of AI-generated content. For testing however, a purely AI-generated dataset is used, since different amount of human samples will be present in the balanced datasets. To prevent over-fitting in the training, we have added dropout layers in the neural network architecture. The LLMs in our dataset are of varied sizes, ranging from 7 billion parameters in Mistral to 1.76 trillion parameters in GPT-4. The AI-generated texts from different LLMs constitute around 20 thousand samples to nearly a million samples. We trained separate models for each of the LLM, and a single model on all the dataset combined.

\begin{table}[!htbp]
\centering
\begin{tabular}{p{0.17\linewidth} | p{0.15\linewidth} | p{0.12\linewidth} | p{0.4\linewidth}}
\hline
\textbf{Model} & \textbf{Params} & \textbf{Size} &\textbf{Source}\\
\hline
Mistral & 7B & 51.5k & \citet{DAIGT-V2}\\
Claude & 130B & 21.8k & \citet{DAIGT-V2}\\
Llama & 70B & 45.4k & \citet{DAIGT-V2}\\
ChatGPT & 175B & 936.8k & \citet{guo2023close} \citet{aaditya_bhat_2023}\\
GPT-4 & 1.76T & 191.5k & \citet{peng2023instruction}\\
Falcon & 180B & 10.3k & \citet{DAIGT-V2}\\
\hline
\end{tabular}
\caption{The number of samples of AI-generated text (a sample is a tuple of 3 sentences) in different datasets used to evaluate our methodology and the size (number of parameters) of the corresponding LLM used to generate the text. \label{tab:LLM}}
\end{table}

\begin{table*}
\centering
\label{tab:model_comparison}
\begin{tabular}{llllllll}
\hline
\textbf{Train/Test} & \textbf{Mistral} & \textbf{Claude} & \textbf{Llama} & \textbf{ChatGPT} & \textbf{GPT-4} & \textbf{Falcon} & \textbf{Average}\\
\hline
Mistral & \color{blue}{98.48} & 74.07 & \color{brown}{71.84} & 64.98 & 76.01 & 73.58 & 76.49\\
Claude & 95.08 & \color{blue}99.63 & 66.93 & 61.19 & 77.89 & 90.34 & 81.85\\
Llama & \color{brown}98.96 & \color{brown}91.12 & \color{blue}99.09 & \color{brown}77.1 & \color{brown}{89.79} & \color{brown}{97.10} & {\bf 92.19}\\
ChatGPT & 80.97 & 34.82 & 49.57 & \color{blue}89.82 & 71.29 & 70.23 & 66.12\\
GPT-4 & 81.90 & 68.68 & 43.07 & 97.69 & \color{blue}99.09 & 70.31 & 76.79\\
Falcon & 92.19 & {89.38} & 69.45 & 69.03 & 74.18 & \color{blue}99.05 & 82.22\\
Combined & 78.18 & 54.41 & 61.63 & 80.46 & 57.74 & 66.83 & 66.54\\
Ensemble & 98.58 & 74.62 & 72.78 & 66.39 & 76.90 & 74.22 & 77.25\\
\hline
\end{tabular}
\caption{Performance comparison of various models on different datasets. Shown are the detection rate in percentage on test sets. The intra-domain performance of models is highlighted in {\color{blue}blue} whereas the best inter-domain performance for a dataset is highlighted in {\color{brown}brown}. For ChatGPT we did not select GPT-4 for best inter-domain performance (even though its detection rate was 97.69\%) as both are from OpenAI and we suspect that the data used for training GPT-4 is inclusive of the data used to train ChatGPT. The average performance achieved (over intra- and inter-domain test sets) is highlighted in bold. \label{tab:genEval}}
\end{table*}

\begin{table*}
\centering
\label{tab:model_ablation}
\begin{tabular}{llllllll}
\hline
\textbf{Train/Test} & \textbf{Mistral} & \textbf{Claude} & \textbf{Llama} & \textbf{ChatGPT} & \textbf{GPT-4} & \textbf{Falcon} & \textbf{Average}\\
\hline
ResNet & {98.77} & 88.63 & \color{brown}{99.52} & 67.06 & 81.76 & 95.92 & 88.61\\
ResNet + ZZ Scheduler & {97.55} & 88.38 & 97.27 & \color{brown}{80.21} & 87.45 & 92.65 & 90.58\\
ZZ ResNet & \color{brown}{98.99} & 87.18 & {98.54} & {77.71} & {87.32} & {95.06} & 90.80\\
ZZ ResNet + ZZ Scheduler & 98.96 & \color{brown}91.12 & 99.09 & 77.1 & \color{brown}{89.79} & \color{brown}{97.10} & {\bf 92.19}\\
\hline
\end{tabular}
\caption{Ablation study on performance of different combinations of ResNet and Scheduler (ref. Algorithm~\ref{alg:algorithm}), on different datasets. Shown are the detection rate in percentage on test sets. The best inter-domain performance for a dataset is highlighted in {\color{brown}brown}. The best average performance achieved (over intra- and inter-domain test sets) is highlighted in bold. ZZ refers to ZigZag in ResNet and Scheduler. \label{tab:genEval_abl}}
\end{table*}

\section{Experimental Results}
\label{experiment}
We evaluated the generalization capabilities and performances of models trained on text data generated by various LLMs. We measure the detection rate (percentage of correctly detecting AI-generated texts) as the performance indicator. Each model is subjected to two different types of evaluations: 
(i) {\it Intra-domain}, and (ii) {\it Inter-domain}. 
In the Intra-domain, a model's performance was evaluated on the same dataset that was used for training. Whereas, when doing Inter-domain evaluation, the model was tested using a test set from a dataset that is not used for the model's training. The results of these evaluations are summarized in Table~\ref{tab:genEval}. To attain a fair comparison of generalization performance, purely AI-generated datasets were used for testing. This is due to the disparity in the sizes of the AI-generated content for different large language models.

The results demonstrate the effectiveness of our proposed model in identifying AI-generated text. Specifically, the model trained on Llama data exhibits robust generalization capabilities, achieving near-perfect performance on Mistral and Falcon datasets. Llama has the best inter-domain performance for all datasets. This indicates the potential of the Llama dataset in training models for AI text detection with high detection rate across various sources.

The inter-domain performance of the models trained on different datasets revealed varying levels of generalization. For example, models trained on Mistral data performed consistently across all the datasets, whereas models trained on ChatGPT data showed poorer generalization capabilities and high variability across datasets.
The model trained on Claude data performs considerably well on Mistral and Falcon data. The model trained on Llama performs great with nearly 92\% detection rate on all datasets except ChatGPT and GPT-4. 
The model trained on GPT-4 has good generalization performance on all datasets except Claude and Llama. The model trained on Falcon has good generalization performance on all datasets except ChatGPT, GPT-4, and Llama.

The combined dataset model performance closely follows the ChatGPT-trained model. This can be attributed to the fact that the ChatGPT dataset accounts for more than 70\% (cf. Table~\ref{tab:LLM}) of the total dataset used to train the combined model. These results show the importance of selecting appropriate training datasets and manifest the impact of dataset-specific characteristics on model performance.

   Even though the Llama data-trained model generalizes better on all the datasets than other models, other models have poor detection rate metrics when testing on Llama data. Contrarily, all models generalize very well and predict Mistral data with great detection rate. Outside of self-datasets, Mistral predicts GPT-4 the best, Claude predicts Mistral the best, Mistral predicts Llama the best, while Llama predicts all other datasets the best. 
   
   Interestingly, Mistral is the smallest model among the models in consideration, in terms of the number of parameters. Our experimental observations highlight the nuances of dataset-specific biases and how they influence the adaptability of different LLMs, suggesting the potential need for more diversified training materials to enhance generalization capabilities across the board.

    Additionally, the novel ZigZag ResNet architecture and the \texttt{ZigZagLROnPlateauRestarts} learning rate scheduler contributed to the enhanced performance of the vision model in classifying text as AI-generated or human-authored. Our \texttt{ZigZagLROnPlateauRestarts} Algorithm~\ref{alg:algorithm} dynamically adjusts the learning rate during the training process of our model based on the plateauing of a performance metric. It increases the learning rate after a specified number of epochs if the performance improves or maintains the current learning rate if the performance does not improve. These results indicate the potential of leveraging image-based techniques for efficient and effective AI text detection.

\section{Ablation Study}
\label{ablation}

        To evaluate the improvement in performance of our ZigZag ResNet and ZigZag Scheduler (ref. Algorithm~\ref{alg:algorithm}), we have conducted an ablation study by training four different models on the best performing dataset (Llama ref. Table~\ref{tab:genEval}). The four experiments are: Vanilla ResNet with no scheduler, Vanilla ResNet with ZigZag Scheduler, ZigZag ResNet with no scheduler, ZigZag ResNet with ZigZag Scheduler. For better comparison in performance, we have used the 5 million parameter version of ResNet-18. Similar to Table~\ref{tab:genEval}, the four trained models are evaluated on all of the six generative AI datasets. Individual performances on each test sets are averaged to get an overall performance score for each of the four models. The model with the best average performance is our ZigZag ResNet along with our ZigZag Scheduler. The model with the weakest performance is the Vanilla ResNet with no scheduler. Our ZigZag Scheduler improves the performance of both the Vanilla ResNet and our ZigZag ResNet. Vanilla ResNet with ZigZag Scheduler is performaning very similar to ZigZag ResNet with no scheduler. Upon analyzing the cells in Table~\ref{tab:genEval}, which are highlighted with brown, we can observe that the ZigZag Architecture has improved the generalization capabilities of the model. Even though, Vanilla ResNet with no scheduler is the best performing model on Llama dataset, the model falters in other datasets. This can be attributed to poor generalization. ZigZag ResNet with ZigZag Scheduler performs the best on three out of the six datasets. While, ZigZag ResNet with no scheduler performs the best on one dataset, even though it is very marginal with its ZigZag scheduler alternative. Surprisingly, Vanilla ResNet with ZigZag Scheduler performs the best on ChatGPT dataset. Overall, our ZigZag ResNet and ZigZag Scheduler offers an improvement in performance of nearly 4\% over Vanilla ResNet-18.

\section{Inference Performance}
\label{scaling}
The goal of our paper is to build a computationally efficient AI-text detection model using convolutional neural networks. The previous section focused on how our model is effective in predicting self as well as other datasets, thereby demonstrating generalization capabilities. In this section, we study the inference performance of our model on GPU (NVIDIA Quadro RTX 8000 GPU) and CPU (Intel Xeon Platinum 8268 CPU @ 2.90GHz). Due to the restriction of the Tensorflow library to switch between CPU and GPU, we created two cloud sessions, one without any GPU, and another with a GPU. Our inference procedure begins with splitting the input text into paragraphs of three sentences. Then, we apply USE on the individual paragraphs and scale them to the range of 0-255. The USE embeddings are then fed to our \texttt{ZigZag ResNet} which internally handles RGB image generation. Finally, we apply sigmoid on the model output to retrieve the probability of a three-sentence paragraph being generated by AI. Finally, all the AI probabilities are averaged to give the overall AI probability for the input text.

We divide the total inference time into preprocessing, forward pass, and output processing time. Preprocessing time constitutes the embedding generation and scaling. Output processing time constitutes sigmoid computation. We have measured inference times for different sizes of input texts (10, 100, 1000, and 10000 sentences) to understand how the model performance scales when running on CPU and GPU. We have also performed tests for different batch sizes (1, 32, 128) to see how the model performance scales with increased batch size.


\begin{figure}[t]
\centering
\includegraphics[width=0.9\columnwidth]{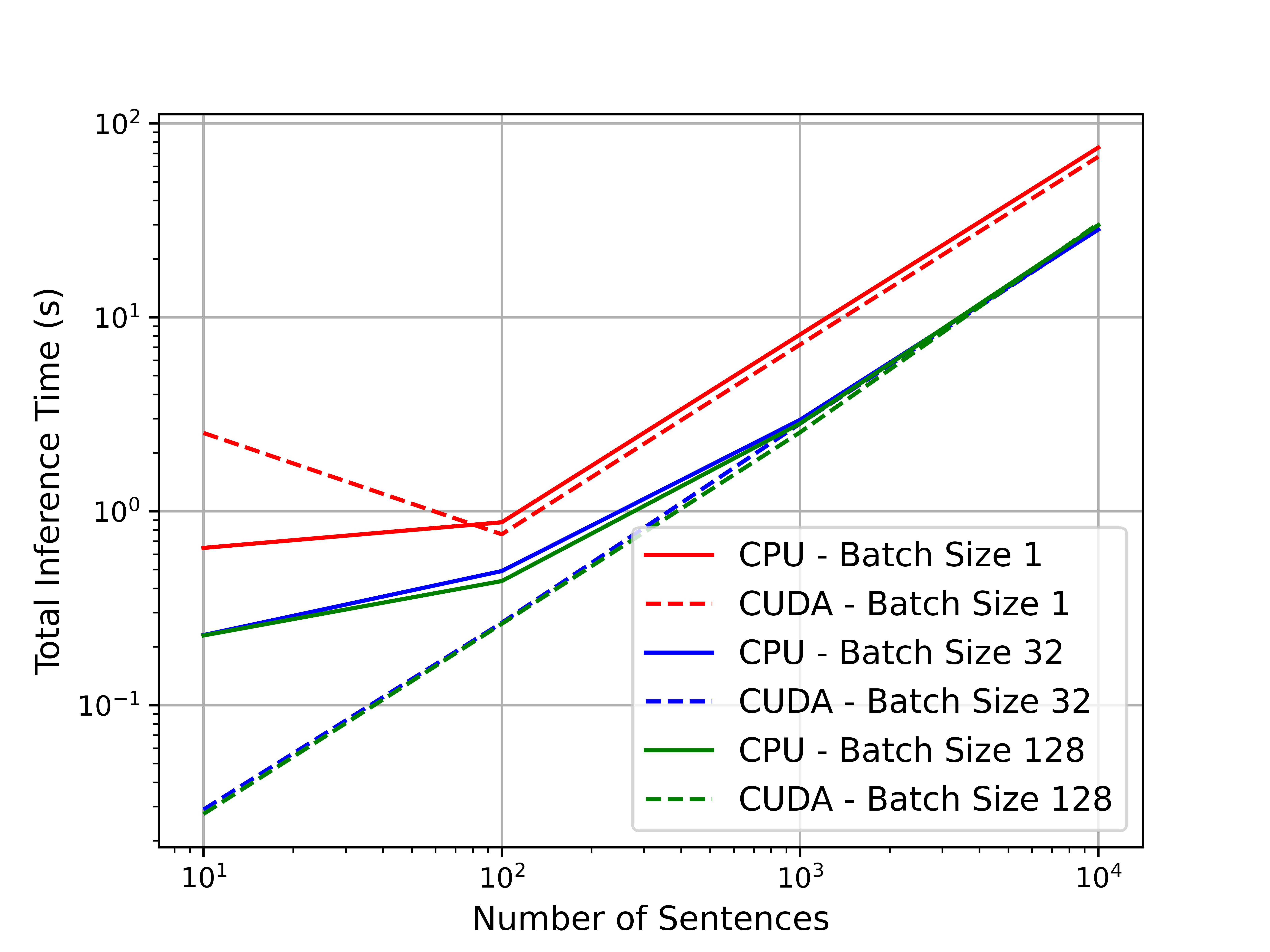} 
\caption{Total Inference Time with respect to number of sentences}
\label{fig5}
\end{figure}

\begin{figure}[t]
\centering
\includegraphics[width=0.9\columnwidth]{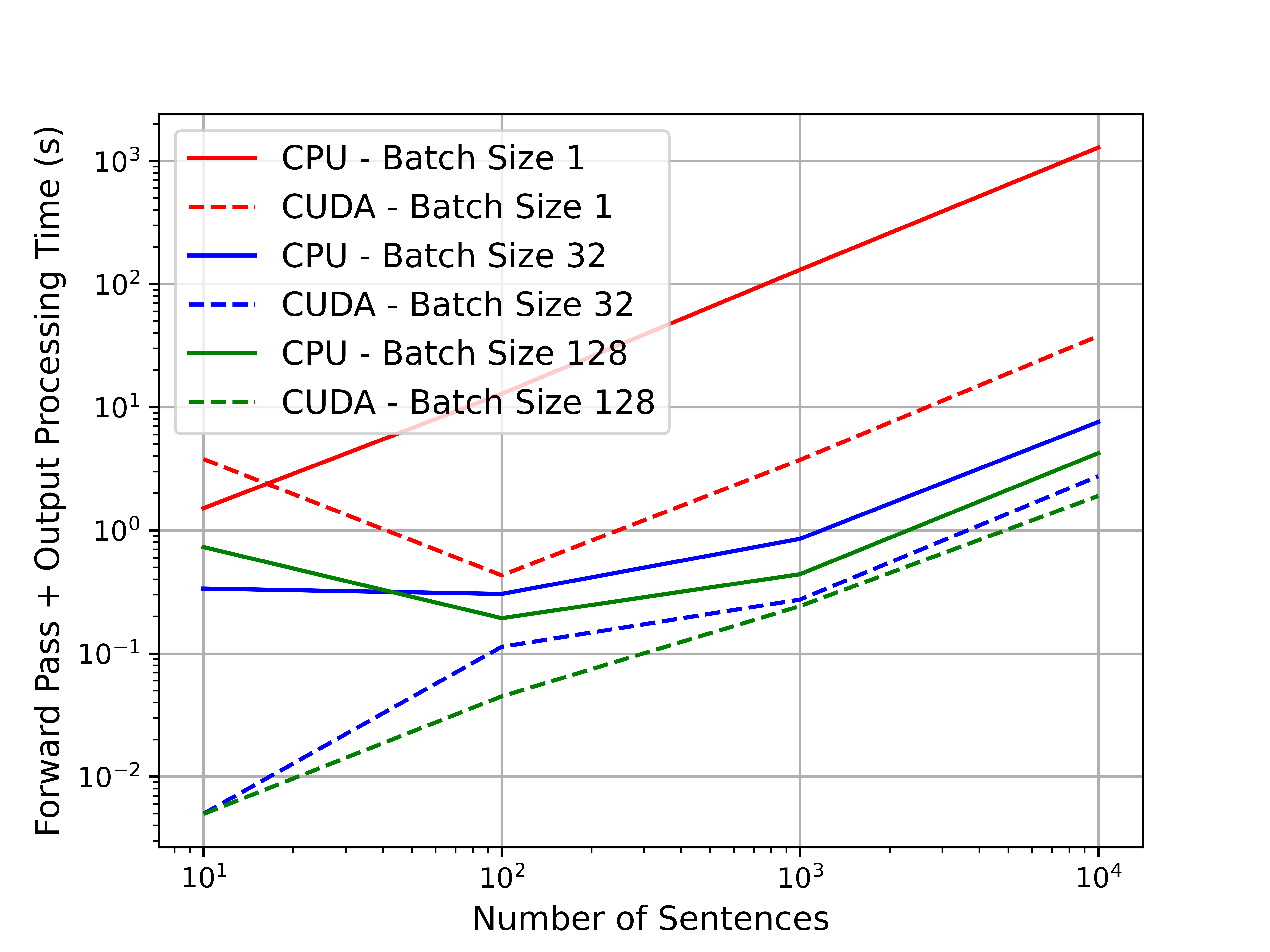} 
\caption{Forward Pass + Output Processing Time with respect to number of sentences}
\label{fig6}
\end{figure}

\begin{figure}[t]
\centering
\includegraphics[width=0.9\columnwidth]{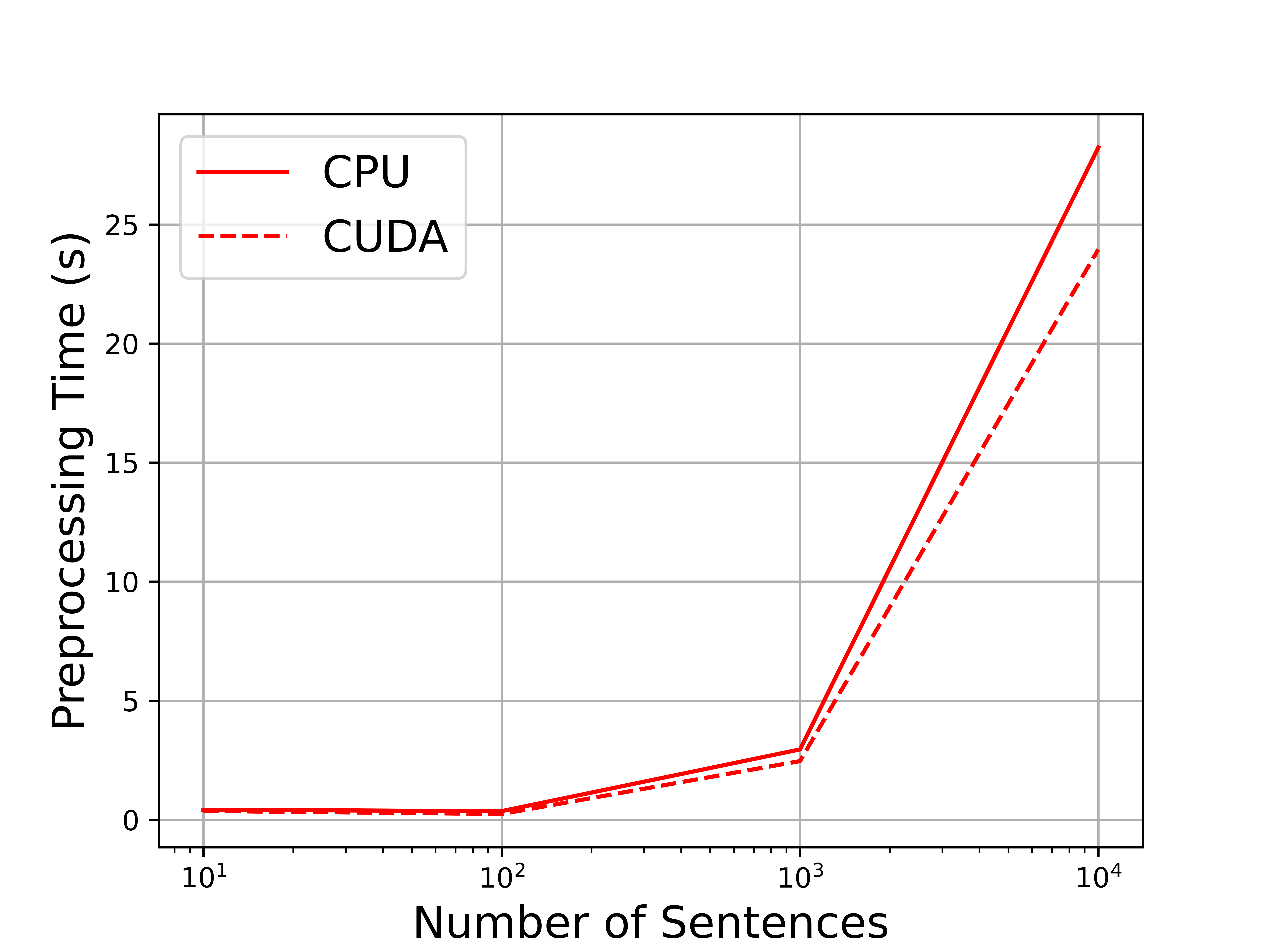} 
\caption{Text Preprocessing Time with respect to number of sentences}
\label{fig7}
\end{figure}

Figure~\ref{fig5} shows that our model scales linearly with an increasing number of sentences. The model that is running on GPU is running considerably better than the one running on CPU for a larger number of sentences. Figure~\ref{fig6} provides a similar insight into the performance, along with the analysis that larger batch sizes lead to smaller forward pass times. Figure~\ref{fig7} shows that the preprocessing time for an increasing number of sentences scales similarly on CPU and GPU. Even though, both CPU and GPU start with similar times for a smaller number of sentences, the difference between them widens as we increase the number of sentences.

\section{Conclusion}
\label{conclusion}
Our work introduces a novel approach for detecting AI-generated text through the conversion of text to image embeddings and the application of vision models, particularly the optimized \texttt{ZigZag ResNet} architecture. The effectiveness of this method has been demonstrated across multiple datasets derived from various Large Language Models, showcasing its potential to enhance the reliability and integrity of academic content. Compared to LLM-based models, our model has lower resource and memory requirements and achieves acceptable (less than ~7.5 ms) end-to-end inference times on both CPU and GPU per three-sentence paragraph.

Our work emphasizes the potential of merging image processing techniques with text analysis to address the challenges posed by the rapid advancement of Generative AI. By continuing to explore and refine these methodologies, we can move closer to ensuring the authenticity and integrity of academic content, ultimately fostering a safer and more trustworthy learning environment. 

\section*{Limitation}
Our coverage of literature may be incomplete
due to the extraordinarily fast pace at which LLMs are evolving and related research is appearing. We have also cited a large number of non-peer-reviewed works, e.g., from arXiv.

Our datasets from different LLMs are of varied sizes, from 10k to around 1M samples. Likewise, when training our models we are restricted by the size of AI-generated content. In addition, we did not find any performance benchmarks related to the datasets we have used. So, it would be difficult to compare how our model performs concerning existing methods.
Our experimental analysis is not comprehensive, missing some ablation studies and other experiments that
could help answer additional questions on the robustness of our methodology. 

We have tested our methodology using only English text. Since we are encoding word embeddings into images, we hypothesize our methodology should work well with other languages, although our hypothesis needs to be tested. When developing a tool based on our methodology, appropriate techniques (paraphrasing input text, adversarial training) need to be employed to have a robust model. 

 Our model for AI text detection makes probabilistic predictions and lacks explainability. Like any machine learning model, our model can generate false alarms and hence it should not used as a primary decision-making tool. At best, its use for plagiarism detection should be restricted to low-impact assignments.

\section*{Ethics Statement}
There are no ethical issues with the data and related resources used in this paper. They are available in open-source and commonly used by other works. The authors have read the ACL Code of Ethics. The fairness of machine learning based models is a well-acknowledged concern. The predictions from our model can be biased and appropriate care should be taken during training (data de-biasing) to ensure fairness and prevent any unintentional harm. 


\bibliography{custom}
\bibliographystyle{acl_natbib}

\end{document}